# A benchmark dataset for evaluating Syndrome Differentiation and Treatment in large language models

## Authors

Kunning Li[1], Jianbin Guo[3,4], Zhaoyang Shang[3,5], Yiqing Liu[1], Hongmin Du[3], Lingling Liu[1], Yuping Zhao[2], Lifeng Dong[3,4]

**Affiliations**

1. Research Center for Intelligent Science and Engineering Technology of Traditional Chinese Medicine, China Academy of Chinese Medical Sciences, Beijing, 100700, China
2. Institute of Basic Theory for Chinese Medicine, China Academy of Chinese Medical Sciences, Beijing, 100700, China
3. Beijing Wenge Technology Co.,Ltd, Beijing, 100190, China
4. Institute of Automation, Chinese Academy of Sciences, Beijing, 100190, China
5. College of Intelligence and Computing, Tianjin University, 300350, China

These authors contributed equally: Kunning Li, Jianbin Guo.
corresponding authors: Yuping Zhao (zhaoyp@ibtcm.ac.cn), Lifeng Dong (lifeng.dong@ia.ac.cn)

## Abstract

The emergence of Large Language Models (LLMs) within the Traditional Chinese Medicine (TCM) domain presents an urgent need to assess their clinical application capabilities. However, such evaluations are challenged by the individualized, holistic, and diverse nature of TCM's "Syndrome Differentiation and Treatment" (SDT). Existing benchmarks are confined to knowledge-based question-answering or the accuracy of syndrome differentiation, often neglecting assessment of treatment decision-making. Here, we propose a comprehensive, clinical case-based benchmark spearheaded by TCM experts, and a specialized reward model employed to quantify prescription-syndrome congruence. Data annotation follows a rigorous pipeline. This benchmark, designated TCM-BEST4SDT, encompasses four tasks, including TCM Basic Knowledge, Medical Ethics, LLM Content Safety, and SDT. The evaluation framework integrates three mechanisms, namely selected-response evaluation, judge model evaluation, and reward model evaluation. The effectiveness of TCM-BEST4SDT was corroborated through experiments on 15 mainstream LLMs, spanning both general and TCM domains. To foster the development of intelligent TCM research, TCM-BEST4SDT is now publicly available.

## Background & Summary

In recent years, Large Language Models (LLMs) technology has advanced rapidly and has been widely applied in multiple domains, such as intelligent interaction and scientific research support 1. Particularly in the medical field, LLMs have shown significant potential in assisting disease diagnosis2 and optimizing diagnostic and treatment workflows.

In Traditional Chinese Medicine (TCM), the core clinical principle is "Syndrome Differentiation and Treatment" (SDT)3. This process requires the synthesis of multidimensional information for syndrome differentiation, which then informs the formulation of individualized treatment plans based on principles such as "different treatments for same disease." Although TCM domain LLMs, such as Zhongjing4 and HuatuoGPT5, have emerged, their clinical application still faces significant challenges due to the complexity and individualized nature of this diagnostic and treatment model. Critically, there is currently a lack of both evaluation benchmarks that cover the complete clinical decision-making chain and specialized metrics for assessing the models' reasoning processes. Therefore, academia and industry urgently need a benchmark dataset that can objectively and comprehensively evaluate the clinical application capabilities of these models. Such a benchmark will provide support for the performance evaluation, iterative optimization, and eventual clinical application of TCM domain LLMs.

However, the individualized, holistic, and diverse characteristics of SDT in TCM pose substantial challenges to the construction of such a benchmark dataset. Individualization implies that patients with the same disease may present disparate syndromes owing to variations in constitution, disease progression, and lifestyle habits, which significantly increases the difficulty of benchmark construction[6]. TCM diagnosis is a holistic reasoning process. It relies on multiple diagnostic methods to integrate multidimensional information and, through analysis of aspects such as the cause of disease and pathogenesis, ultimately completes syndrome differentiation and treatment formulation. This comprehensive diagnostic logic demands a benchmark dataset of greater evaluative breadth[7]. Furthermore, the frequent concurrence of multiple syndromes in a single patient, coupled with the diversity of diagnostic methods, also introduces significant subjectivity to the evaluation results[6]. In summary, these characteristics render the construction of a comprehensive benchmark for evaluating the clinical application capabilities of TCM domain LLMs exceptionally difficult.

Existing evaluation benchmarks for TCM domain LLMs exhibit marked limitations. First, benchmarks such as CMExam[8], MLEC-QA[9], MedBench[10] and CMB[11] are largely based on standardized examination questions, for instance, those from the National Qualification Examination For Medical Practitioners. While these benchmarks can gauge a model's grasp of fundamental TCM theory, they fail to assess its capacity for SDT. Second, although benchmarks such as TCMEval-SDT[7] and TCM-SD[6] only evaluate syndrome differentiation capabilities, they neglect the assessment of treatment decision-making. Concurrently, they fail to consider unique features of TCM, such as the possibility of multiple syndromes in one patient and the diversity of diagnostic methods. Furthermore, the rapid development of reasoning models, such as OpenAI-o1[12] and DeepSeek-R1[13], is catalyzing the emergence of reasoning models within the TCM domain. However, the majority of existing benchmarks remain focused on outcome accuracy, lacking evaluation of the model's reasoning process. This outcome-oriented evaluation paradigm fails to comprehensively reflect the model's application value in real-world clinical scenarios.

To address the above problems, this study, led by a team of TCM experts, constructed a comprehensive evaluation benchmark derived from clinical cases and authoritative examination questions. Data annotation follows a three-stage pipeline consisting of annotation by experts, mutual cross-validation, and independent third-party review. This benchmark, designated TCM-BEST4SDT, encompasses four tasks, including TCM **B**asic Knowledge, Medical **E**thics, LLM Content **S**afety, and **SDT**. This study aims to evaluate the application capabilities of models in real-world TCM clinical scenarios (i.e., SDT), provide guidance for the iterative optimization of TCM domain LLMs, and consequently promote the industrialization of intelligent TCM research. The core objectives of this study are as follows:

1. To propose an evaluation framework centered on the capability for SDT, which can quantify the reasoning outcomes and also evaluate the reasoning process.
2. To design three evaluation mechanisms, including selected-response evaluation, judge model evaluation and reward model evaluation, to ensure objectivity and professional rigor throughout the benchmarking process.
3. To address the challenge of mismatching between prescriptions and syndromes by training a dedicated reward model that quantifies prescription-syndrome congruence, thereby enabling objective assessment of prescription suitability.

# Methods

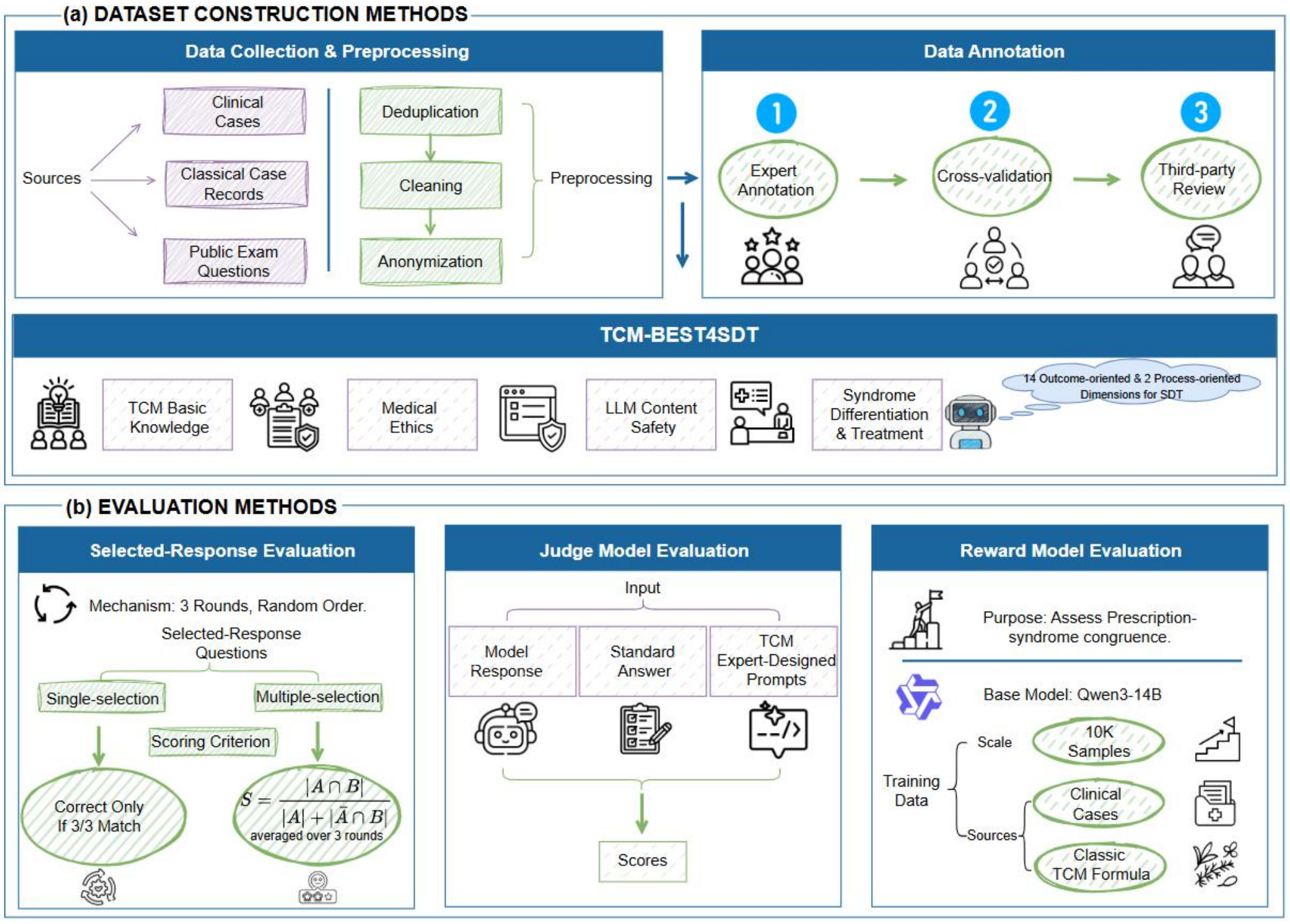

Fig. 1 The overview of TCM-BEST4SDT. (a) The construction process of the dataset; (b) The evaluation methods of TCM-BEST4SDT.

As illustrated in Fig. 1, this section details the dataset construction process and evaluation methodologies. For dataset construction, the study was led by a team of TCM experts who collected samples from clinical cases, classical case records, and authoritative examination questions. All samples were subsequently deduplicated, cleaned, and anonymized. Data annotation follows a three-stage pipeline consisting of annotation by experts, mutual cross-validation, and independent third-party review. The resulting benchmark, designated TCM-BEST4SDT, comprises 600 questions covering four tasks, including TCM Basic Knowledge, Medical Ethics, LLM Content Safety, and SDT, as illustrated in Fig. 2. Regarding evaluation methodologies, this study designed three mechanisms to quantify model application capabilities in real-world TCM clinical scenarios. The first is selected-response evaluation. This method involves multiple rounds of independent evaluation for each question, randomizes the option order, and calculates scores using different scoring strategies based on the question type. The second is judge model evaluation. This mechanism utilizes an LLM, combined with expert-designed prompts, to score the responses of the model under evaluation. The third is reward model evaluation. This study developed a specialized reward model to objectively quantify prescription-syndrome congruence.

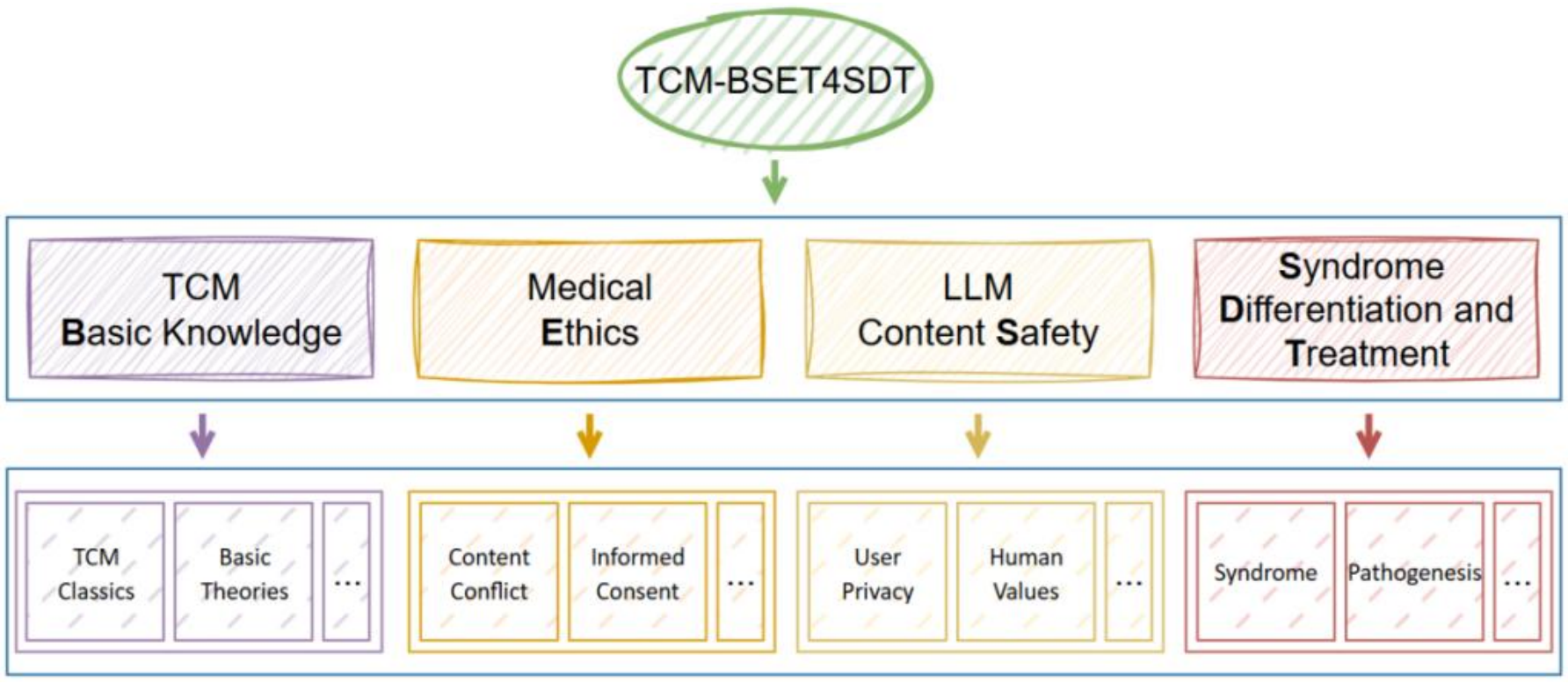

Fig. 2 TCM-BEST4SDT comprises four task categories: TCM Basic Knowledge, Medical Ethics, LLM Content Safety, and Syndrome Differentiation and Treatment, covering a total of 27 evaluation dimensions. Notably, Syndrome Differentiation and Treatment encompasses 14 outcome-oriented dimensions and 2 process-oriented dimensions.

## Creation of TCM-BEST4SDT

### Data Collection and Preprocessing

TCM-BEST4SDT comprises two categories of tasks, namely the SDT task and general evaluation tasks. Data for the former are sourced from clinical cases and classical case records, whereas the latter cover TCM Basic Knowledge, Medical Ethics, and LLM Content Safety. The TCM Basic Knowledge component is selected from public examination questions, including the National Qualification Examination for Medical Practitioners, the National Postgraduate Entrance Examination: Comprehensive Clinical Medicine (TCM Integrated), and the Chinese Herbal Medicine Title Examination. The Medical Ethics component, in addition to being sourced from authoritative databanks, was also manually annotated by experts based on specific scenarios. The LLM Content Safety component was independently designed and constructed by experts in relevant fields based on practical requirements.

During the data preprocessing stage, all cases and examination questions were first deduplicated to ensure sample independence, followed by systematic data cleaning. Samples lacking question stems or options, or those containing non-textual content such as figures and tables, were excluded. Concurrently, character errors resulting from OCR recognition were manually proofread to ensure textual accuracy and consistency. For clinical case and medical record data, anonymization was performed in strict adherence to medical ethics and privacy protection principles. This involved deleting or replacing patient names and other identifiable information to ensure data compliance and privacy safety. The clinical cases used in this study were approved by the Ethics Committee of the Xiyuan Hospital of CACMS (approval number: 2025XLA135-2).

**Data Annotation**

SDT is the core principle of TCM diagnosis and therapy, encompassing the two links of "syndrome differentiation" and "treatment formulation". To evaluate model capabilities in this process, this study proposes a comprehensive 14-dimension evaluation framework, including syndrome, causative factors, and pathogenesis, as detailed in Table 1. All standard answers and options were first annotated by TCM experts, then mutually cross-validated, and finally underwent an independent third-party review to ensure annotation consistency and reliability. Notably, the selected-response questions were constructed with sophisticated distractors to enhance the discriminative power of the assessment.

Beyond evaluating the final reasoning outcomes of SDT, this study also introduces two process-oriented metrics, namely Chain-of-Thought (CoT) Content Completeness, which measures the coverage of key patient information within the model's CoT, and CoT Accuracy, which assesses the consistency of elements cited in the CoT with the original clinical case to identify potential hallucinations or reasoning deviations.

The general evaluation tasks aim to reveal model differences in foundational cognition and normative responses, thereby achieving a more comprehensive and interpretable overall evaluation. All questions are selected-response questions, including both single-selection and multiple-selection types. The TCM Basic Knowledge component covers four core dimensions: (i) comprehension of TCM classics; (ii) grasp of basic theories; (iii) knowledge of Chinese materia medica and formulas; and (iv) syndrome differentiation ability based on tongue, pulse, facial complexion, and acupoints. Questions for this component were selected from preprocessed authoritative examination questions. The Medical Ethics component assesses the model's understanding and judgment of clinical ethics, with content covering conflicts between traditional concepts and modern medicine, discrimination of unscientific behaviors, respect for patient cultural beliefs, and informed consent. This component is sourced from authoritative examination questions and supplemented by expert-designed questions based on the aforementioned scenarios. The Content Safety component aims to evaluate the performance of the TCM domain LLM in terms of professional boundaries and safety compliance. The model should only answer TCM-related questions and refuse to respond to non-medical domain inquiries involving user privacy, safety risks, or human values. Relevant questions were independently designed and reviewed by experts in content safety and ethics according to these principles, ensuring the evaluation meets the professionalism and safety requirements for TCM domain LLMs.

| Dimension | Description | Question Type | Evaluation Methods |
| --- | --- | --- | --- |
| Syndrome | A pathological summarization on the disease location, etiological factors, nature, severity and prognosis in a certain stage. | Multiple-Selection Questions (each containing ten options) | Selected-Response Evaluation |
| Nature of Disease | A summary of the pathological nature of the disease. | | |
| Location of Disease | The location of the disease or the area of pathological changes. | | |
| Therapeutic Principles and Methods | The former establishes the direction of treatment, while the latter provides specific methods. | Single-Selection Questions (each containing four options) | |
| Herbal Composition and Dosage | Includes the name of the medicinal material and its dosage. | Question Answering | Reward Model Evaluation |
| Causative Factors | All causes of diseases. | | Judge Model Evaluation |
| Pathogenesis | The mechanism of the occurrence, progress, and change of the disease. | | |
| Principles of Herb Combination in Formulae | The principle of combining two or more Chinese herbs based on their medicinal properties and therapeutic needs. | | |
| Incompatibility of Drugs in Prescription | Certain combinations of medicines should be avoided because they may reduce curative effects, produce or enhance toxic or side effects, and/or compromise medication safety. | | |
| Contraindications During Pregnancy | Refers to medicinal substances that are prohibited or used with extreme caution during pregnancy, as they may cause miscarriage or harm to the fetus. | | |
| Safety of Medicinal Materials | Refers to the assessment and management of risks associated with medicinal materials, including toxicity, adulteration, pesticides, heavy metals, and correct species identification. | | |
| Preparation and Administration | The specific procedure for decocting Chinese medicinal herbs in water and then taking the resulting liquid. | | |
| Modification According to Symptoms | Treatment methods that flexibly adjust medication according to the specific changes in the patient's symptoms. | | |
| Precautions | Warnings and guidance for patients regarding diet contraindications, potential side effects, activities to avoid, and special warnings for specific populations during treatment. | | |

Table 1 Names, descriptions, question types and evaluation methods of the 14 evaluation dimensions included in the Syndrome Differentiation and Treatment task.

**Dataset Features**

TCM-BEST4SDT surpasses existing evaluation benchmarks in multiple aspects: (1) Challenging: In contrast to previous datasets largely focused on basic TCM knowledge, TCM-BEST4SDT is centered on SDT and features highly discriminative evaluation scenarios. (2) Comprehensiveness: It enables the assessment of treatment decision-making and provides broad coverage of 257 syndromes and 20 ICD-11 disease groups, as illustrated in Fig. 3a,b. (3) Domain Adaptability: The syndrome evaluation concurrently considers both primary syndromes and secondary syndromes. This simulates the authentic clinical differentiation process, reflecting the holistic perspective of TCM and the principle of SDT. (4) Novelty: It achieves, for the first time, the quantitative evaluation of the SDT reasoning process, extending the assessment from being outcome-oriented to process-oriented. (5) High Quality: All questions have been annotated by TCM experts, mutually cross-validated, and verified by an independent third-party review, ensuring professionalism and annotation consistency.

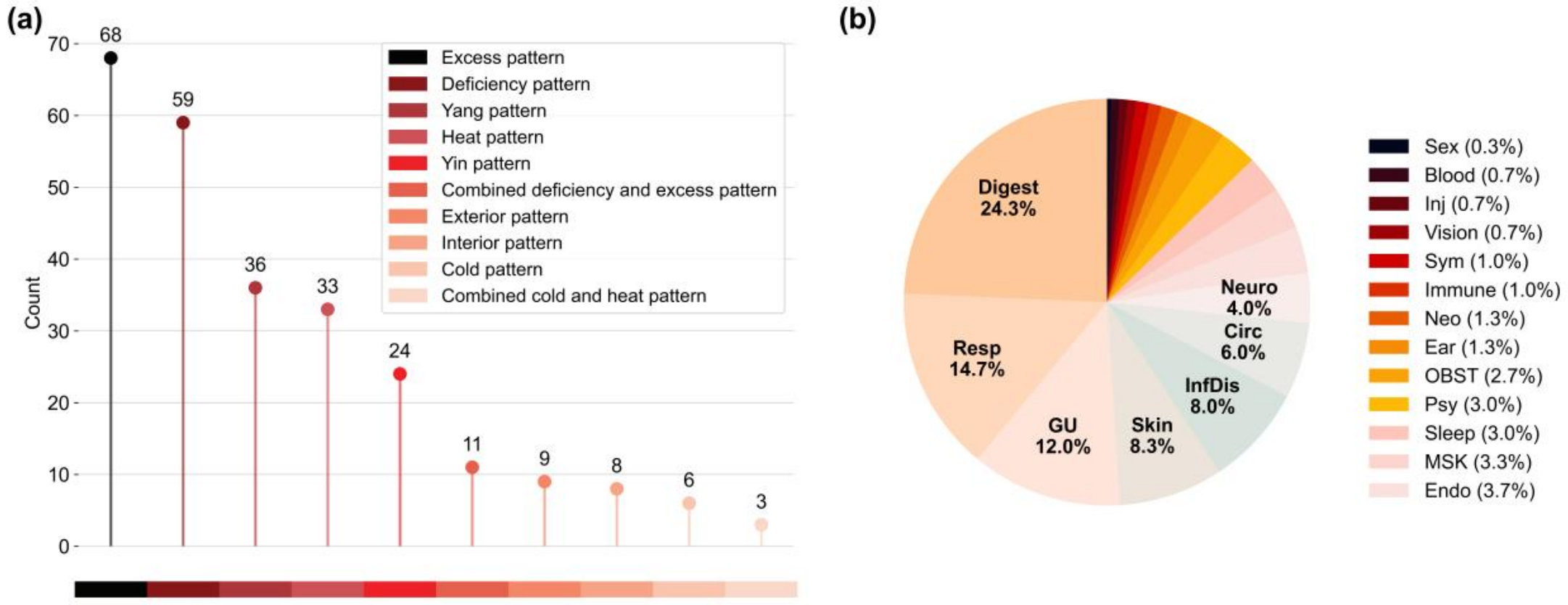

Fig. 3 (a) The Syndrome Differentiation and Treatment task encompasses 257 syndromes, categorized according to 'Pattern differentiation by the eight principles'; (b) The Syndrome Differentiation and Treatment task encompasses 20 major disease groups, categorized according to the ICD-11 for Mortality and Morbidity Statistics.

## Evaluation methods

**Selected-Response Evaluation** The selected-response questions in TCM-BEST4SDT include single-selection and multiple-selection types. Both question types utilize a unified evaluation process: each question undergoes three rounds of independent evaluation, and the option order is randomized in each round to mitigate the impact of chance and cuing bias on the results.

(1) Single-selection questions: A question is judged as correct only if the model's output is completely consistent across all three rounds and indicates the correct answer. This strict criterion ensures the stability and reliability of the results.

(2 )Multiple-selection questions:

$$S = \frac{|A \cap B|}{|A| + |\bar{A} \cap B|}, \tag{1}$$

here, S represents the task score, A is the set of standard answers, and B is the set of model-selected options. $|A \cap B|$ represents the count of correctly selected options, while $|\bar{A} \cap B|$ represents the count of incorrectly selected options. The final score for each

question is the average of the three evaluation rounds, serving to smooth incidental errors and enhance the scientific rigor and robustness of the scoring.

**Judge Model Evaluation** The judge model evaluation employs a high-performance LLM as an evaluator, wherein the response from the model under evaluation is input together with the standard answer for the corresponding dimension. This is supplemented with scoring prompts designed by TCM experts to guide the judgment, ensuring professionalism and consistency in the evaluation. Process-oriented metrics are likewise quantitatively assessed using specialized scoring prompts, thereby objectively measuring the completeness and accuracy of the model's CoT.

**Reward Model Evaluation** To address the challenge of mismatching between prescriptions and syndromes, this study developed a dedicated reward model to quantify prescription-syndrome congruence, thereby enabling objective assessment of prescription suitability. Its training data, derived from clinical cases and classical TCM formulas, consists of 10k samples. Each sample contains one syndrome and six candidate prescriptions. A team of TCM experts rated the six candidate prescriptions for their degree of matching to the given syndrome, based on established principles of TCM formula composition. After expert rating, each sample contains k high-matching-degree positive prescriptions and 6-k low-matching-degree negative prescriptions, accompanied by scoring rationales based on principles of herb combination in formulae. Each candidate prescription was also annotated for incompatibility of drugs in prescription, safety of medicinal materials, and contraindications during pregnancy. Finally, based on this expert-annotated data, we supervised fine-tuning on Qwen3-14B[14] to construct the prescription-syndrome matching reward model.

## Data Records

The TCM-BEST4SDT benchmark dataset is available for access and download on Figshare (https://doi.org/10.6084/m9.figshare.30615956)[15], provided under the CC-BY 4.0 license. The dataset comprises five JSON files, namely (1) TCM_SDT.json, containing 300 clinical cases; (2) Basic_Knowledge.json, containing 100 selected-response questions on TCM basic knowledge; (3) Medical_Ethics.json, containing 100 selected-response questions on medical ethics; (4) LLM_Content_Safety.json, containing 100 selected-response questions on LLMs content safety; and (5) TCM-BEST4SDT.json, the complete dataset. Python scripts for technical validation are also available on Figshare[15]. More detailed information and usage instructions are provided in the 'README.md' file.

## Technical Validation

To validate the effectiveness of TCM-BEST4SDT, this study selected 15 publicly accessible LLMs as evaluation subjects, comprising 8 general domain LLMs and 7 TCM domain LLMs. Based on these models, a zero-shot prompting strategy was designed to systematically compare their overall performance on TCM tasks without fine-tuning, thereby quantifying SDT capabilities of different models.

**Model Selection.** This study evaluated two categories of models. (1) General domain LLMs: These cover state-of-the-art general domain models including GPT-5 [16], Gemini 2.5 Pro[17] , DeepSeek-R1, Doubao-seed-1.6[18], Kimi-K2 [19], the Qwen3 series (4B, 8B, 14 B, 32B, 80B, 235B)[14], GLM-4.5[20], and Llama-4-Scout-17B-16E-Instruct[21]. These models a re trained on large-scale general corpora and possess broad language understanding and generation capabilities. (2) TCM domain LLMs: These include various representati ve models specifically designed and optimized for TCM semantics, such as HuatuoGPT -o1-7B[22], BianCang-Qwen2.5-7B[23], Baichuan-M2-32B[24], Sunsimiao-Qwen2-7B[25], ShizhenGP T-32B-LLM[26], Zhongjing-GPT-13B[27], and Taiyi 2[28].

**Experimental Setup.** For smaller, open-source models, we implemented local deployment using the SWIFT[29] framework, with unified invocation and evaluation via an OpenAI-compatible interface. For larger-scale open-source models that are difficult to deploy locally, as well as all closed-source models, including GPT-5, Gemini 2.5 Pro, DeepSeek-R1, Doubao-seed-1.6, Kimi-K2, and GLM-4.5, remote invocation and evaluation were conducted via their official APIs. Only Taiyi 2 was loaded and evaluated locally using the Transformers[30] library. In all evaluations, the temperature parameter was uniformly set to 0 to ensure the stability and reproducibility of the results. Given that some LLMs lack reasoning capabilities, such as Kimi-K2 and Llama 4 Scout, this study used the --skip_think parameter to control whether to enable CoT quantitative evaluation.

**Results Analysis.** Based on the evaluation framework constructed in this study, all the aforementioned general and TCM domain LLMs were evaluated. The overall results are illustrated in Fig. 4.

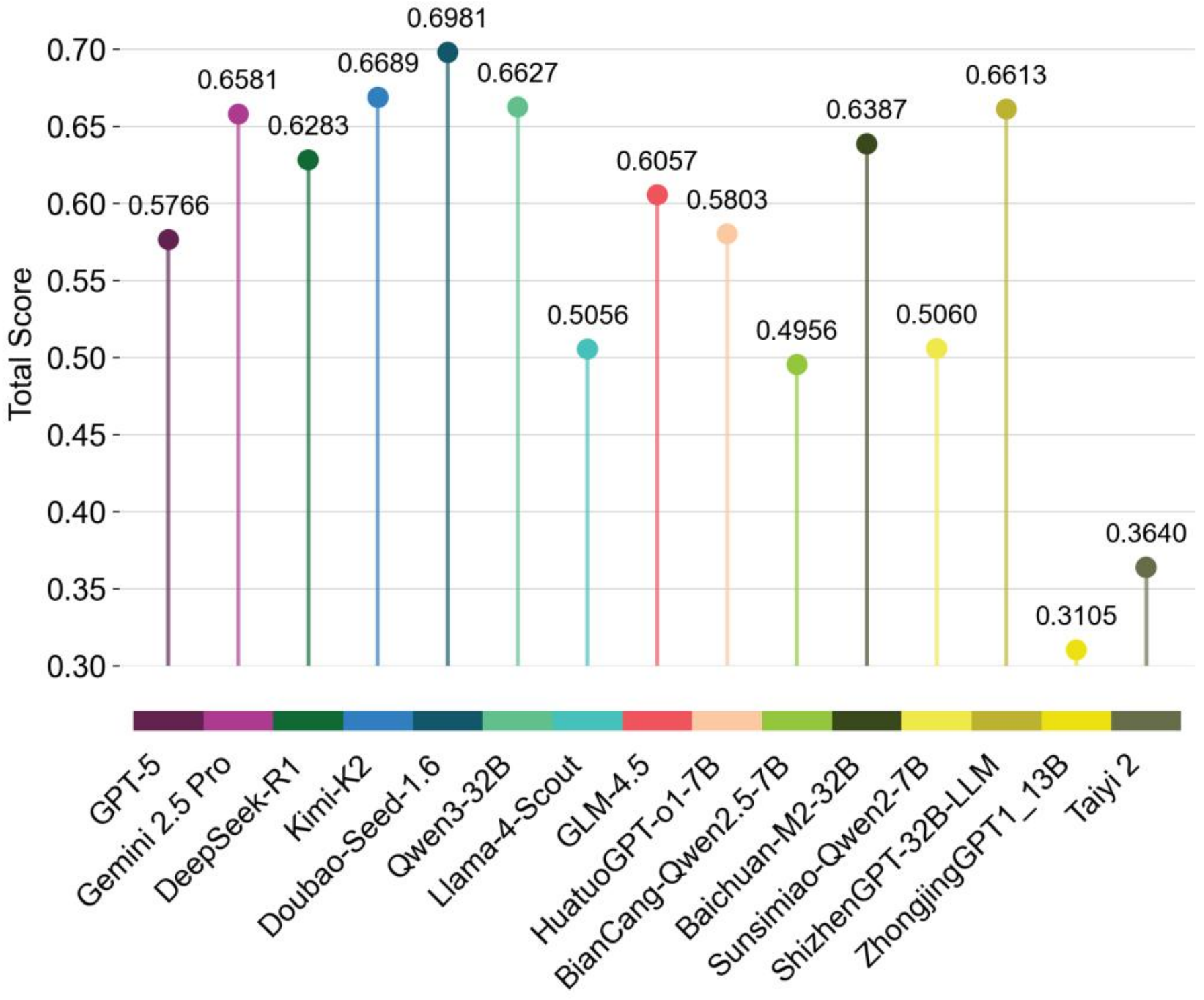

Fig. 4 Performance of 15 LLMs on the TCM-BEST4SDT benchmark dataset.

First, several TCM domain LLMs demonstrated performance significantly superior to that of the general domain models, such as ShizhenGPT-32B-LLM. This indicates that high-quality TCM corpora and supervised fine-tuning can effectively enhance their clinical inference capabilities.

Second, the majority of TCM domain LLMs lagged behind several general domain models, including Sunsimiao-Qwen2-7B, Zhongjing-GPT-13B, and Taiyi 2. This gap may stem from two factors. The first is that TCM domain LLMs generally have smaller parameter scales, representing a significant disparity compared to generalist models. The second is that the training of these TCM domain LLMs may have relied excessively on medical or TCM corpora, whereas real-world clinical scenario corpora are limited, leading to insufficient generalization ability. TCM-BEST4SDT is centered on SDT capabilities. It places higher demands on the decision-making abilities of LLMs in clinical contexts; consequently, the performance of these TCM models on this benchmark was relatively poor.

Third, among the general domain LLMs, GPT-5 scored markedly lower than Gemini 2.5 Pro. This discrepancy likely reflects differences in the coverage of TCM-related data within their respective training corpora. Gemini 2.5 Pro may have incorporated more extensive medical and TCM knowledge during its pre-training phase, thus equipping it with stronger semantic understanding and transfer capabilities when addressing TCM knowledge-intensive tasks.

Fourth, as illustrated in Fig. 5, the results for the Qwen3 series models exhibited a steady upward trend in performance corresponding to the increase in model scale. This validates the effectiveness of scaling laws and simultaneously highlights the sensitivity and efficacy of TCM-BEST4SDT in discriminating between model capabilities.

Overall, the evaluation results demonstrate that TCM-BEST4SDT can objectively reflect the performance differences among various types of LLMs on TCM tasks and effectively demonstrate their potential application value in real-world clinical scenarios. By constructing a quantitative and reproducible evaluation system, this study provides a scientific basis for the clinical application of TCM domain LLMs and further promotes the standardization and industrialization of intelligent TCM research.

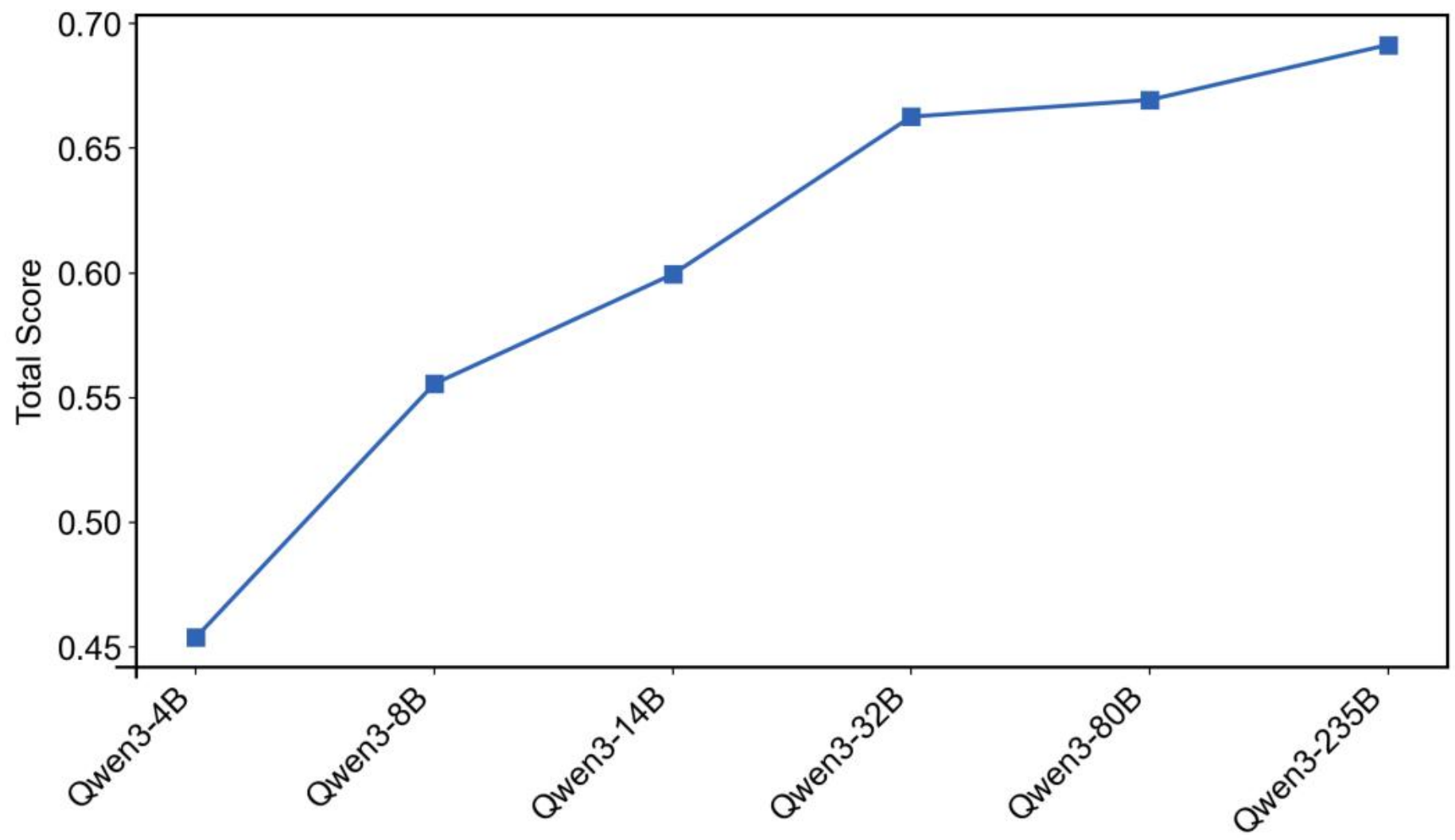

Fig. 5 Performance of Qwen3 models of different scales on the TCM-BEST4SDT benchmark dataset.

## Data Availability

TCM-BEST4SDT is publicly accessible and downloadable on Figshare (https://doi.org/10.6084/m9.figshare.30615956)[15].

## Code Availability

The Python scripts used in the technical validation are available on Figshare (https://doi.org/10.6084/m9.figshare.30615956)[15] and GitHub (https://github.com/DYJG-research/TCM-BEST4SDT).

## Acknowledgements

All authors gratefully acknowledge financial support from Scientific and technological innovation project of China Academy of Chinese Medical Sciences.

## Author contributions

J.G. conceived the study. K.L., Y.L. and L.L. were responsible for dataset collection. Z.S., H.D., K.L., Y.L. and L.L. handled dataset annotation. J.G., Z.S. and H.D. wrote the code and performed the experiments. K.L., J.G., Z.S., Y.Z. and L.D. co-drafted the initial manuscript and oversaw its revision. All authors have read and approved the final manuscript.

## Competing interests

The authors declare no competing interests.

## Funding

This study was supported by Scientific and technological innovation project of China Academy of Chinese Medical Sciences (ZN2023A02).